%% file: main.tex
\documentclass[letterpaper]{article}
\input{preamble}
\title{\papertitle}
\input{authors}
\begin{document}
\maketitle
\input{sections/body}
\clearpage
\input{sections/ethics}
\bibliography{references}
\end{document}

%% file: preamble.tex
\usepackage[preprint]{aaai2027}
\usepackage{graphicx}
\usepackage{amsmath,amssymb}
\usepackage{booktabs}
\usepackage{float}
\usepackage{natbib}
\usepackage{url}
\newcommand{\papertitle}{Worth Another Look:\\Constrained World-Model Planning for Visual Inspection}
\newcommand{\wm}{\textsc{WM}}
\newcommand{\rr}{\textsc{RR}}
\newcommand{\fp}{\textsc{FP}}
\newcommand{\stopact}{\textsc{stop}}
\newcommand{\calA}{\mathcal{A}}
\newcommand{\calC}{\mathcal{C}}
\newcommand{\calD}{\mathcal{D}}
\newcommand{\calM}{\mathcal{M}}
\floatstyle{ruled}
\newfloat{algorithm}{tbp}{loa}
\floatname{algorithm}{Algorithm}

%% file: authors.tex
\author{
    \mbox{Ziyue Kang},
    \mbox{Zixi Qi},
    \mbox{Yuchen Dai},
    \mbox{Nana Lv},
    \mbox{Xirong Cheng},\\
    \mbox{Liang Liu},
    \mbox{Ying Li},
    \mbox{Haoyan Wei},
    \mbox{Haiyang Yao\thanks{Corresponding author (\texttt{yaohaiyang@sust.edu.cn}).}}
}
\affiliations{
    Shaanxi University of Science and Technology, Xi'an, Shaanxi, China\\
    yaohaiyang@sust.edu.cn
}

%% file: sections/body.tex
\begin{abstract}
Visual inspection in field robotics and environmental monitoring must localize multiple targets under strict compute and sensing caps. Exhaustive sweeps waste calls on absent classes, while myopic and search-free policies stumble over heterogeneous tariff cliffs or discard viewport edits whose value emerges only after re-querying. We formulate inspection over frozen detectors as neuro-symbolic observation planning under partial observability, decoupling a compact recurrent world model (\wm) from exact symbolic feasibility and tariff rules. Receding-horizon beam search with online replanning jointly schedules class queries, crops, enhancements, and termination (\stopact), paired with slack-aware shadow pricing and legal fallback at larger budgets. Across eight terrestrial, optical, and sonar benchmarks, multi-step lookahead outperforms static schedules and search-free offline RL under tight budgets---gaining up to 28.4 AP percentage points over static priority on RUOD and 15.1 on ARIS---while dynamic pricing lifts large-budget performance on all eight domains at 11\,ms planning cost. Ultimately, pairing lightweight learned dynamics with exact symbolic constraints turns frozen detectors into adaptive, budget-aware inspection agents without retraining.
\end{abstract}

\section{Introduction}
Field robotics, marine debris monitoring, and sonar inspection must localize multiple targets under strict latency and compute budgets~\cite{hong2020trashcan,liu2021duo,fu2023ruod,valdenegro2025marine}. Open-vocabulary detectors~\cite{liu2023grounding}, enhancements, and adaptive visual imagination~\cite{yu2026when} raise coverage in degraded scenes, yet exhaustive sweeps or static cascades violate compute caps and waste calls on absent classes. Classical active localization~\cite{caicedo2015active} only refines single-object boxes via homogeneous transforms---it cannot allocate shared budgets across heterogeneous detector tiers, crops, enhancements, and termination ($\stopact$).

\begin{figure*}[t]
\centering
\includegraphics[width=\textwidth]{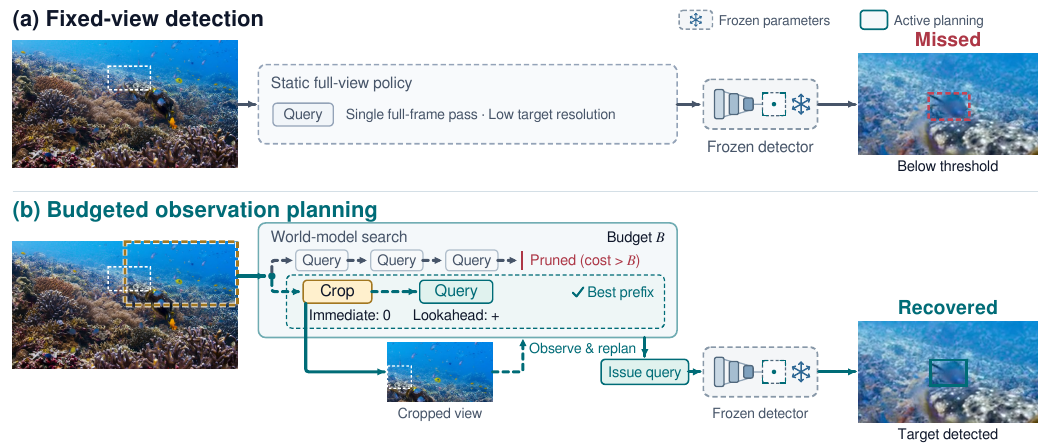}
\caption{\textbf{Static full-view detection versus budgeted observation planning} using the same frozen detector. \textbf{(a)}~A static single-step full-frame query suffers from low effective target resolution, leaving the small underwater target below the confidence threshold. \textbf{(b)}~Our receding-horizon world-model search evaluates multi-step action prefixes under exact budget and feasibility masks, pruning over-budget query sequences while selecting a zero-immediate-reward viewport crop whose predicted lookahead gain is realized after replanning and querying the cropped view.}
\label{fig:motivation}
\label{fig:teaser}
\end{figure*}

Principled allocation points to partially observable Markov decision process (POMDP) planning~\cite{kaelbling1998planning,silver2010monte} with value-of-information branching~\cite{laouar2026leveraging}, offline RL~\cite{kumar2020conservative}, and latent world models~\cite{hafner2019learning,xi2024neuro}. Directly applying these paradigms over frozen detectors hits three pathologies: (1)~in the \emph{delayed-reward viewport trap}, full-frame passes miss small targets below threshold (Figure~\ref{fig:teaser}(a)), while quadrant crops earn zero immediate reward ($r_t=0$) and are pruned by greedy heuristics before gains appear upon re-querying (Figure~\ref{fig:teaser}(b)); (2)~under tight caps ($B=2$), \emph{heterogeneous tariff cliffs} arise when an alternative call with premium $\kappa>1$ over a fast query ($\kappa=1$) leaves $b_1<1$, stranding a second call that TD backups in conservative Q-learning (CQL)~\cite{kumar2020conservative} fail to resolve; and (3)~\emph{dual conservatism and horizon trade-offs} trigger early stopping under fixed penalties~\cite{stocco2024addressing} while deeper rollouts compound model error~\cite{nguyen2024lookahead}.

Rather than forcing neural networks to approximate hard constraints from visual traces~\cite{xi2024neuro}, we decouple a compact recurrent world model (\wm, $247\text{K}$--$268\text{K}$ parameters) from exact symbolic feasibility and tariff rules. External masks prune infeasible prefixes~\cite{allen2026simplifying} and enforce budget, geometry, deduplication, and edit-cap invariants outside latent transitions---letting \wm{} specialize in multi-step query utility. At $B=2$, raising $H=1$ to $H=3$ boosts $\text{AP}_{50:95}$ (abbreviated AP) from $0.345$ to $0.404$ on Real-world Underwater Object Detection (RUOD) and $0.350$ to $0.392$ on Detecting Underwater Objects (DUO), with adaptive gains up to $+28.4$ AP points over fixed priority on RUOD and $+15.1$ on Acoustic Resolution Imaging Sonar (ARIS). At $B=8$, slack-aware shadow pricing and legal fallback curb early stopping, lifting AP across all eight datasets at $11\,$ms planning latency.

\noindent In summary, our contributions are threefold:
\begin{itemize}
    \item \textbf{Neuro-Symbolic Sensing Formulation:} Casting multi-class inspection over frozen detectors as a costed POMDP decoupling compact dynamics from exact symbolic legality and tariff rules.
    \item \textbf{Lookahead with Budget-Aware Pricing:} Resolving zero-reward edit bridges and tariff cliffs via constrained beam search, while curbing high-budget stopping via dynamic pricing and legal fallback.
    \item \textbf{Cross-Modal Mechanistic Analysis:} Disentangling static priors, scene-adaptive targeting, rollout fidelity, and beam scaling across eight terrestrial, optical, and sonar tasks.
\end{itemize}

\section{Related Work}
\paragraph{Sensing and partial observability.}
Belief-space POMDPs~\citep{kaelbling1998planning}, POMCP~\citep{silver2010monte}, and value-of-information branching~\citep{laouar2026leveraging} maintain state distributions to select informative observations. Leaving ground-truth labels and unqueried outputs hidden under deterministic perception, we pair a compact GRU history encoding $z_t$ with an exact symbolic control tuple $q_t$.

\paragraph{Active visual perception.}
Active localization~\citep{caicedo2015active} tracks single targets via homogeneous box transforms, while adaptive test-time scaling gates generative visual imagination~\citep{yu2026when}. Instead, we schedule frozen open-vocabulary~\citep{liu2023grounding} and fine-tuned detectors, crops, enhancements, and termination ($\stopact$) across a multi-class vocabulary under latency-calibrated tariffs.

\paragraph{Planning with learned models.}
Latent dynamics~\citep{chua2018deep,hafner2019learning} and lifted action learners like ROSAME~\citep{xi2024neuro} infer transition rules from trajectories. Supplying exact symbolic preconditions and relevance pruning~\citep{allen2026simplifying} externally lets our $250\text{K}$-parameter world model focus strictly on query utility, isolating lookahead against discrete CQL~\citep{kumar2020conservative} and return-ranked imitation.

\paragraph{Constraints and search horizon.}
Recursive dual ascent curbs conservatism under fixed POMDP cost penalties~\citep{stocco2024addressing}, while Upper Confidence Bounds for Trees (UCT) exhibits nonmonotonic horizon trade-offs under model error~\citep{nguyen2024lookahead}, motivating our joint study of search depth ($H\in\{1,3,5\}$), beam width ($K$), and zero-retrain shadow pricing.

\section{Constrained Observation Planning}
\subsection{Task, Actions, and Information}
Each episode supplies an image $I$, an ordered vocabulary $\calC$ of $C$ classes, and budget $B$. The target configuration $Y$ is hidden. The controller encodes the history $h_t$ of tool outputs and image statistics in $z_t$ and maintains the exact control state
\begin{equation}
 q_t=(b_t,v_t,\mathcal{H}_t,n_t,\eta_t),
\end{equation}
tracking remaining budget $b_t$, viewport and edit path $v_t$, query history $\mathcal{H}_t$, step count $n_t$, and consecutive edit count $\eta_t$.

\begin{figure*}[t]
\centering
\includegraphics[width=\textwidth]{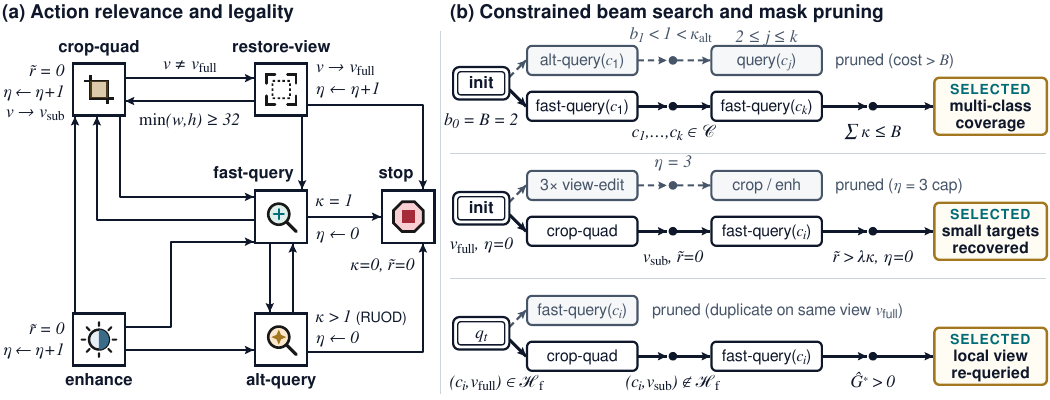}
\caption{\textbf{(a) Planar observation action relevance and legality graph.} An edge $a_1\to a_2$ shows a possible continuation under $\calM$ after $\mathrm{StepControl}(q_t,a_1)$. Nodes give the tariff $\kappa$, planning reward $\tilde r$, view $v$, and edit count $\eta$. \textbf{(b) Constrained beam search and mask pruning on state--action DAGs.} Solid paths are selected prefixes. At $B=2$ on RUOD, two fast class queries fit ($\sum\kappa\le B$). An alternative leaves $b_1<1$. A zero-reward crop can enable a later detection. A fourth consecutive edit is masked at $\eta=3$. Fast-query history $\mathcal H_{\mathrm f}$ blocks same-view repeats. The local re-query lies on the best retained prefix ($\widehat G^*>0$).}
\label{fig:legality_dag}\label{fig:method_graph}
\end{figure*}

The $2C+8$ actions comprise fast and alternative queries for each class $c\in\calC$---modeling target-conditioned tool routing and class-specific query invocation under a unified costed interface---along with four quadrant crops, full-view restoration, two domain-specific enhancements, and $\stopact$. Each query runs on the current viewport. Accepted boxes are projected to image coordinates and merged by class-wise non-maximum suppression (NMS) at intersection-over-union (IoU) $0.5$. A crop doubles the effective resolution of its quadrant. Enhancements adjust contrast, gamma, or white balance.

The observation $o_t\in\mathbb{R}^D$, $D=178+3(|\calA|-10)$, includes 33 thumbnail quality features, seven detection summaries, viewport and control indicators, class query and discovery masks, action counts, and legality masks. A $4\times4$ grid records box count, mean confidence, and area occupancy per cell. Ground truth $Y$ is used only for offline reward labels and held-out evaluation.

\subsection{Costs, Reward, and Feasibility}
Each dataset fixes action tariffs $\kappa(a)$ from median synchronized training-split step latencies, normalized to $\kappa(\mathrm{fast})=1$. We cap total charges at $B\in\{2,4,6,8\}$ and measure wall time including search and state updates.

Let $F(\calD_t,Y;\calC)$ be macro-averaged per-class F1 at IoU $0.5$ for accumulated detections $\calD_t$. Its change supplies the offline reward:
\begin{equation}
 r_t=F(\calD_{t+1},Y;\calC)-F(\calD_t,Y;\calC).
 \label{eq:reward}
\end{equation}
Only queries return boxes, so edits and $\stopact$ yield $r_t=0$. On held-out episodes, $\text{AP}_{50:95}$ is class-averaged COCO mean AP over IoU thresholds $0.50{:}0.05{:}0.95$, evaluated for the target vocabulary. We also report micro recall at IoU $0.5$ and total-count mean absolute error (MAE).

As formalized in Algorithm~\ref{alg:inspection} and illustrated by the planar legality graph in Figure~\ref{fig:legality_dag}(a), the mask $\calM(q_t,a)$ requires $\kappa(a)\le b_t$ and $n_t<T=8$ for nonterminal actions. A crop requires both dimensions to be at least 32 pixels. Restoration requires a cropped or enhanced view ($v_t\ne v_{\mathrm{full}}$). Edits require $\eta_t<3$, and enhancement is masked on already enhanced views. A query must satisfy $(a,v_t)\notin\mathcal H_t$, where $a$ identifies the tool and class. Successful queries reset $\eta_t$. Zero-cost $\stopact$ is always legal, so $\sum_t\kappa(a_t)\le B$.

\section[A Learned Model with Explicit Constrained Search]{A Learned Model with\\*Explicit Constrained Search}
\subsection{History Encoding and Prediction}
A 128-dimensional gated recurrent unit (GRU) encodes the four most recent observations and actions. Observations are standardized by training statistics, concatenated with a 24-dimensional action embedding, and projected through a 128-unit sigmoid linear unit (SiLU) layer. A GRUCell $f_\theta(z_t,a)$ predicts latent transitions; a two-layer SiLU head predicts reward through $\tanh$ and cost through $\mathrm{softplus}$; a decoder reconstructs normalized observations. Depending on vocabulary size, \wm{} contains 246,952--267,622 parameters.

All learned controllers use the same offline trajectories from six behavior policies: random, round robin, enhance--query, fixed schedule, heuristic, and alternative-query priority. The world-model loss combines one-step reward regression, smooth-$L_1$ cost regression, latent consistency, and discounted three-step reconstruction over 88 perceptual dimensions (33 quality, 7 detection, 48 grid features). Rewards are supervised on observed prefixes. Planning supplies declared tariffs and zero edit rewards directly, with exact control transitions outside the learned dynamics.

\subsection{Receding-Horizon Beam Search}
Algorithm~\ref{alg:inspection} couples learned latent rollouts with exact symbolic control procedures ($\mathrm{Legal}$ and $\mathrm{StepControl}$) to search feasible action prefixes $\pi=(a_0,\ldots,a_{L-1})$, $L\le\min(H,T-n)$, ranking each prefix by its predicted net return:
\begin{equation}
 \widehat G(\pi\mid h_t,q_t)=
 \sum_{d=0}^{L-1}\gamma^d\big[\tilde r(\widehat z_d,a_d)-\lambda_t \kappa(a_d)\big],
 \label{eq:search}
\end{equation}
where $\widehat z_0=z_t$, $\gamma=0.95$, $\lambda_t=\lambda_0=0.02$ in the original comparisons, and $\tilde r(\widehat z_d,a_d)=\widehat r_\theta(\widehat z_d,a_d)\cdot\mathbb{I}[a_d\in\calA_{\mathrm{query}}]$.

\input{sections/algorithm}

Over $H$ depths, batched expansion scores $O(HK|\calA|)$ candidates and retains up to $K=24$ feasible prefixes per depth. Zero-cost $\stopact$ self-loops carry terminated prefixes forward. Let $\widehat G^*$ be the best retained prefix score. The controller executes $a_t=\pi^*_0$ when $\widehat G^*>0$ and replans, or takes $\stopact$ otherwise. As shown in Figure~\ref{fig:legality_dag}(b), with positive price $\lambda_t$, at $H=1$ any edit $e$ scores $-\lambda_t \kappa(e)<0$ against $0$ for $\stopact$ and is excluded, whereas at $H\ge 2$ an edit--query pair $(e,a)$ yields net return
\begin{equation}
 \gamma\widehat r_\theta(f_\theta(z_t,e),a)
 -\lambda_t[\kappa(e)+\gamma \kappa(a)],
 \label{eq:edit-query}
\end{equation}
enabling search to schedule crops or enhancements when downstream query gain outweighs combined costs and to preserve budget across multi-query sequences.

\paragraph{Budget-aware pricing and stopping.}
To curb conservative termination when remaining budget slack is large, we evaluate a budget-aware \emph{Dynamic} schedule using the slack-adaptive shadow price
\begin{equation}
 \lambda_t=\lambda_0\max\left\{0,1-\frac{\max(0,b_t-1)}{\max(1,T-n_t)}\right\},
 \label{eq:shadow}
\end{equation}
where $\mathrm{Price}(q_t)$ returns $\lambda_t$ under Dynamic mode and $\lambda_0$ under the fixed-price baseline, held constant within each candidate prefix and updated after every real step. Let $\mathcal U(q_t)$ denote legal fast queries for classes absent from the successful query history $\mathcal{H}_t$. When search proposes $\stopact$ while $b_t\ge1$ and $\mathcal U(q_t)\ne\emptyset$, the fallback selects $\arg\max_{a\in\mathcal U(q_t)}\widehat r_\theta(z_t,a)$. Both mechanisms operate purely at test time over unchanged frozen checkpoints ($\lambda_0=0.02$).

\section{Experimental Protocol}
\suppressfloats[t]
\subsection{Datasets and split control}
Table~\ref{tab:data} spans three sensing domains: terrestrial Common Objects in Context (COCO-5) and Trash Annotations in Context (TACO-4) \citep{lin2014microsoft,proenca2020taco}; underwater optical DUO, RUOD, and TrashCan (16-class Material split) \citep{liu2021duo,fu2023ruod,hong2020trashcan}; and underwater acoustic sonar benchmarks from Acoustic Resolution Imaging Sonar (ARIS), Side-Scan Sonar Target Detection (SCTD), and Underwater Acoustic Target Detection (UATD) \citep{valdenegro2025marine,zhang2022self,yang2022uatd}. To prevent frame-level or cross-dataset leakage, all splits enforce perceptual-hash near-duplicate removal and whole-sequence partitioning prior to training (detailed in the supplement). Dedicated controllers are trained per benchmark under a unified architecture to evaluate planner generality across optical and acoustic sensing modalities.

\begin{table}[b]
\centering
\small
\begin{tabular}{lrrrr}
\toprule
Dataset & $C$ & Train & Dev & Test\\
\midrule
COCO-5 & 5 & 2,048 & 256 & 500\\
TACO-4 & 4 & 1,050 & 225 & 209\\
DUO & 4 & 5,671 & 256 & 1,111\\
RUOD & 10 & 6,433 & 1,135 & 3,296\\
TrashCan & 16 & 4,937 & 1,118 & 1,157\\
ARIS & 10 & 1,665 & 50 & 153\\
SCTD & 3 & 257 & 46 & 54\\
UATD & 10 & 1,210 & 206 & 1,600\\
\bottomrule
\end{tabular}
\caption{Curated controller-training, development, and held-out test splits across terrestrial, underwater optical, and sonar benchmarks, with $C$ target classes per dataset.}
\label{tab:data}
\end{table}

\paragraph{Frozen tools and training.}
Fast tools are frozen YOLO11s detectors (input 640, confidence 0.25): pretrained on COCO for COCO-5 and fine-tuned on the respective training splits elsewhere. Alternative terrestrial queries use Grounding DINO Tiny; underwater optical and sonar queries use enhancement fusion. Controllers share history encoding, normalization, trajectories, and AdamW optimization ($10^{-3}$ learning rate, batch 128, up to 50 epochs, development patience 10), with seeds 42, 2024, and 3407.

\paragraph{Comparators.}
We evaluate eleven online methods under identical legality masks: four fixed schedules---round robin (\rr), enhance--query (Enh), confidence-triggered cascade (Cascade), and a fixed crop/query template (Fixed)---along with uniform random legal actions (Random), two search-free learned controllers---return-ranked imitation (Imit.) and discrete CQL---myopic reward-only greedy selection (RO), and our world-model receding-horizon beam search (\wm) at horizons $H\in\{1,3,5\}$ (denoted $\wm_H$ or \wm{} ($H$)). Imitation trains a policy head on the highest-return logged continuation, while CQL learns long-horizon Q-values via temporal-difference (TD) backups and a conservative regularizer to represent model-free value planning. We also construct an offline fixed-priority reference (\fp) that queries classes in decreasing order of training-set initial F1 gain, yielding 1,056 live seed--method--test evaluations plus deterministic static reference curves.

\paragraph{Measurement and fairness.}
We evaluate AP via COCOeval over accumulated, NMS-merged detections in original-image coordinates alongside micro recall (IoU $0.5$) and per-image total-count MAE, reporting means and sample standard deviations (SD) across three controller seeds together with seed-paired differences. Every query targets an explicit class under frozen, latency-calibrated tariffs. Serial wall-clock runtime---covering resident tool execution, image decoding, feature extraction, beam search, and diagnostic logging---is profiled on a single RTX 5070 Ti GPU (16,303~MiB; PyTorch 2.13.0+cu130; Ultralytics 8.4.204) using uncached live tools across the first 32 paired images per dataset and seed, while all accuracy metrics span full held-out test sets.

\FloatBarrier
\section{Results and Analysis}

\begin{figure*}[!t]
\centering
\includegraphics[width=\textwidth]{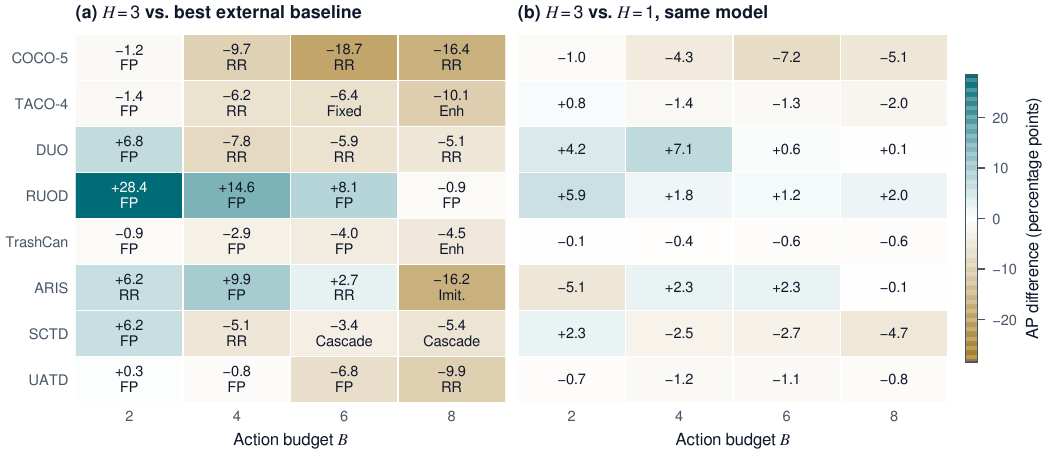}
\caption{Test $\text{AP}_{50:95}$ differences with fixed $\lambda_0=0.02$ in percentage points at the same action budget. Panel (a) subtracts the best external baseline from $H=3$. Cells name one maximizing policy, with ties labeled in the order listed in the text (RR first). FP denotes fixed priority, Enh enhance--query, and Imit.\ imitation. Panel (b) subtracts $H=1$ from $H=3$ for the same model, paired by training seed.}
\label{fig:horizon}
\end{figure*}

\begin{table*}[!t]
\centering
\small
\setlength{\tabcolsep}{0pt}
\input{tables/budget2_compact}
\caption{Test $\text{AP}_{50:95}$ at $B=2$ across all twelve methods and eight datasets (mean $\pm$ sample SD over three training seeds for learned controllers; deterministic schedules have zero seed variance; bold marks the highest mean per column). At $B=2$, $H=3$ and $H=5$ yield identical metrics because any two queries ($\kappa\ge 1$) plus at most one intermediate edit exhaust the allowance within three active decisions. Fixed priority ($*$) is an offline deterministic reference across all eight benchmarks ($n=1$ source seed on COCO-5, TACO-4, and DUO; three identical repeats elsewhere). Full four-budget tables appear in the supplement.}
\label{tab:budget2}
\end{table*}

\subsection{Where Lookahead Improves Detection}
Table~\ref{tab:budget2} reports the complete tight-budget comparison. Three-step planning exceeds the strongest external baseline on five of eight datasets (Figure~\ref{fig:horizon}(a)); Figure~\ref{fig:horizon}(b) isolates $H=3$ versus $H=1$ under identical weights. On RUOD, its margins over fixed priority are 28.4, 14.6, and 8.1 AP percentage points (pp) at $B=2,4,6$. At $B=2$, margins over the best external baseline are 6.8~pp on DUO, 6.2~pp on SCTD, 6.2~pp on ARIS, and 0.3~pp on UATD, while fixed priority leads $H=3$ by 1.2~pp on COCO-5 and 1.4~pp on TACO-4. Paired depth gains are $0.059\pm0.011$ AP on RUOD (1.669 versus 1.393 charged units) and $0.042\pm0.019$ on DUO. DUO's three-step policy also exceeds reward-only by 2.8~pp at matched cost (1.977 versus 1.978).

A world-model variant leads overall on five datasets: three-step planning on DUO, RUOD, and SCTD, and one-step planning on ARIS ($0.279$ AP versus $0.228$ at $H=3$) and UATD ($0.075$ versus $0.068$). Reward-only leads on COCO-5 ($0.188$), while fixed priority leads on TACO-4 ($0.172$) and TrashCan ($0.033$). RUOD and ARIS retain three-step advantages over external baselines at $B=4,6$; larger budgets expose conservative stopping.

\paragraph{Recovery and counting.}
At $B=2$, three-step RUOD recall rises from 0.088 to 0.544 over round robin and count MAE falls from 4.398 to 2.232. TrashCan instead loses recall (0.190 to 0.133) and increases MAE (1.246 to 1.370). ARIS improves AP with unchanged recall (0.347) and MAE (0.725). DUO's AP advantage over reward-only accompanies slightly worse recall and count error.

\begin{figure*}[!t]
\centering
\includegraphics[width=\textwidth]{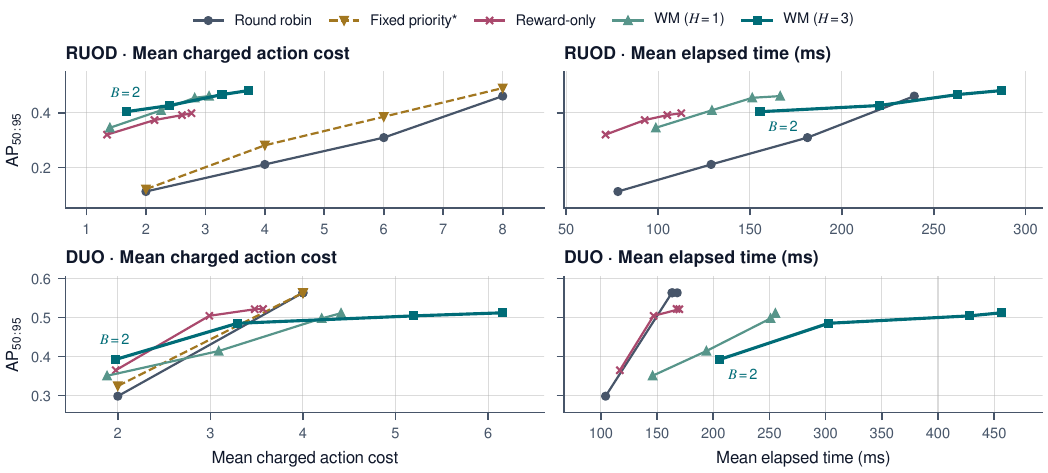}
\caption{Charged cost and elapsed time versus $\text{AP}_{50:95}$ on RUOD and DUO. AP uses full test sets; live uncached timings use the same 32 frozen images per dataset and three seeds, in one GPU process. Dynamic combines the shadow price and legal fast-query fallback. Fixed priority has action costs only.}
\label{fig:resource_curves}
\end{figure*}

Figure~\ref{fig:resource_curves} places these AP gains against charged action cost and elapsed time; both resources are analyzed below.

\subsection{How the Planner Allocates Queries}
A static class order can improve AP even when every image receives the same queries. Let $\Delta_{\mathrm{AP}}=\mathrm{AP}_{\wm_3}-\mathrm{AP}_{\rr}$ denote the gain of three-step planning over round robin at a fixed budget. Inserting the fixed-priority reference (\fp) gives
\begin{equation}
 \Delta_{\mathrm{AP}}
 =(\mathrm{AP}_{\fp}-\mathrm{AP}_{\rr})
 +(\mathrm{AP}_{\wm_3}-\mathrm{AP}_{\fp}).
 \label{eq:decomposition}
\end{equation}
The first term on the right measures the gain from adopting the train-ranked class order. The second compares the planner with that order and includes differences in class selection, detector choice, viewport edits, and stopping.

\begin{figure}[t]
\centering
\includegraphics[width=\columnwidth]{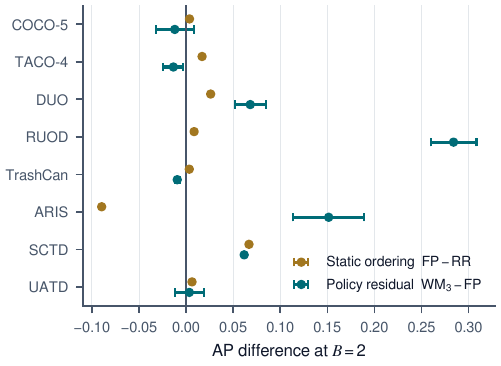}
\caption{Decomposition of tight-budget ($B=2$) AP gains into static class-priority ordering ($\fp-\rr$) and the adaptive planning residual ($\wm_3-\fp$, mean $\pm$ paired seed SD).}
\label{fig:gains}
\end{figure}

For RUOD at $B=2$, the static term is $+0.009$ AP and the adaptive policy residual is $+0.284\pm0.024$ (Figure~\ref{fig:gains}). DUO exhibits a static gain of $+0.026$ and an adaptive residual of $+0.068\pm0.016$. On the sonar benchmarks, three-step planning yields positive adaptive residuals of $+0.151\pm0.038$ AP on ARIS and $+0.062\pm0.001$ on SCTD, with $+0.003\pm0.015$ on UATD, whereas static priority remains ahead on 16-class TrashCan ($0.009\pm0.002$) and the terrestrial splits COCO-5 ($0.012\pm0.020$, static $+0.004$) and TACO-4 ($0.014\pm0.011$, static $+0.017$). Pairing $H=3$ against $H=1$ under identical weights further isolates the contribution of lookahead depth from static prior ordering.

\paragraph{Scene-specific selection.}
To test whether those choices align with individual scenes, we match query traces to held-out annotations offline. Let $Q_{ic}=1$ if class $c$ was queried at least once on image $i$ and $Y_{ic}=1$ if it has a non-crowd instance there. The observed match rate is the fraction of queried image--class pairs for which the class is present. The reference uniformly permutes whole query sets among the same $N$ images. This preserves the mix of queried classes and the distribution of queries per image, and breaks their correspondence to a particular scene. The difference from the exact permutation expectation is
\begin{equation}
 \Delta_{\rm match}=
 \frac{\sum_{i,c}Q_{ic}Y_{ic}-\frac{1}{N}\sum_c Q_{\cdot c}Y_{\cdot c}}
      {\sum_{i,c}Q_{ic}},
 \label{eq:match}
\end{equation}
where dots denote sums over images. We also compute \emph{present-class coverage}, $\sum_{i,c}Q_{ic}Y_{ic}/\sum_{i,c}Y_{ic}$, measuring the fraction of truly present classes that receive at least one query. Evaluating both statistics across the 96,960 logged test episodes quantifies scene-adaptive targeting directly from recorded decisions.

Table~\ref{tab:query_behavior} shows positive scene-matching lift at both horizons across all eight datasets. On RUOD, $H=3$ gives $+34.3\pm2.4$~pp lift ($57.9\%$ match versus $23.6\%$ under permutation), expands present-class coverage from $51.7\%$ to $58.5\%$, and changes the query set on $39.9\%$ of images. It recovers more instances on $15.2\%$ of images and fewer on $3.9\%$. DUO coverage rises from $72.2\%$ to $76.2\%$, with more recovery on $10.3\%$ and less on $2.9\%$. ARIS changes query sets on $51.4\%$ of scenes but trades breadth ($43.8\%$ coverage versus $49.8\%$) for edited-view inspection.

\begin{table*}[!t]
\centering
\small
\setlength{\tabcolsep}{0pt}
\input{tables/query_behavior}
\caption{Query behavior: original rows cover 96,960 test episodes; Dynamic rows add 24,240 episodes at $B=8$. Entries are three-seed means, with scene lift in pp $\pm$ sample SD. Coverage counts present image--class pairs. Edited-view and stop-reserve rates count episodes; reserve requires voluntary STOP with at least one fast unit left. Dynamic's AP difference is against fixed-price $H=3$.}
\label{tab:query_behavior}
\end{table*}

\paragraph{Which images account for extra detections?}
Subtracting matched instances under $H=1$ from $H=3$ per image, over $97\%$ of net extra instance detections on RUOD and DUO at $B=2$ occur on images where three-step lookahead alters the queried class set---holding in every training seed and confirming that tight-budget lookahead improves AP primarily by broadening scene-relevant class discovery.

\paragraph{Preserving room for a second query.}
A fast query costs $1.000$ unit versus $1.014$ on RUOD and $1.184$ on DUO for an alternative query. At $B=2$, calling the alternative tool first leaves less than $1.000$ unit ($0.986$ or $0.816$), precluding a second detector call. Whereas $H=1$ chooses the alternative query on $35.9\%$ of RUOD and $10.9\%$ of DUO episodes, $H=3$ lookahead avoids that bottleneck ($0.3\%$ on RUOD, $0.0\%$ on DUO), lifting two-query completion from $40.9\%$ to $68.0\%$ on RUOD and $86.4\%$ to $97.6\%$ on DUO.

\paragraph{Where edits enter.}
At $B=2$, edited-view queries appear on $19.6\%$ of ARIS, $12.7\%$ of UATD, and $8.4\%$ of TrashCan episodes (below $0.2\%$ on RUOD/DUO). At $B=8$, $H=3$ shifts toward multi-step refinement, querying edited views on $83.4\%$ of DUO, $77.8\%$ of SCTD, $72.2\%$ of COCO-5, $63.4\%$ of ARIS, and $45.2\%$ of RUOD episodes (Table~\ref{tab:query_behavior}).

\subsection{Budget Use, Stopping, and Planning Overhead}
At $B=8$ under fixed $\lambda_0=0.02$, voluntary early termination with an affordable fast query remaining (\emph{Stop reserve} in Table~\ref{tab:query_behavior}) occurs on $86.5\%$ of TrashCan, $73.9\%$ of UATD, and $65.7\%$ of RUOD episodes. Enabling the Dynamic schedule drives premature affordable-query stopping to zero on RUOD, TrashCan, ARIS, and UATD while sharply expanding present-class coverage ($19.6\%$ to $54.1\%$ on TrashCan, $40.8\%$ to $57.5\%$ on UATD, and $69.3\%$ to $78.4\%$ on RUOD). Mean AP improves across all eight datasets by $+0.6$ to $+6.7$~pp---raising TrashCan from $0.019$ to $0.036$, UATD from $0.089$ to $0.116$, ARIS from $0.368$ to $0.427$, COCO-5 from $0.224$ to $0.291$, and RUOD from $0.481$ to $0.507$---shifting termination onto the eight-step horizon ($86.1\%$--$100\%$ reach $n_t=8$ without issuing $\stopact$).

\paragraph{Sensing charges versus wall-clock latency.}
Figure~\ref{fig:resource_curves} separates charged costs from serial runtime on RUOD and DUO. On RUOD at $B=8$, Dynamic reaches $0.507$ AP at $6.573$ units, beating both fixed-price $H=3$ ($0.481$ at $3.724$) and full-budget fixed priority ($0.491$ at $8.000$), and improving paired RUOD AP at all four caps ($+1.12, +1.83, +2.05, +2.59$~pp). On four-class DUO, round robin covers the vocabulary in four full-view steps ($0.563$ AP at $4.000$ units), whereas $H=3$ and Dynamic commit $83.4\%$ of large-budget episodes to localized edits within the shared eight-step horizon ($0.518$ AP at $6.956$ units).

\begin{table}[t]
\centering
\small
\setlength{\tabcolsep}{0pt}
\input{tables/timing_breakdown}
\caption{Matched serial GPU runtime at $B=2$ (three-seed means; 32 paired images per dataset; live uncached tools). Top: stage times in ms. Bottom: beam sweep at $H=3$, with full-test $\text{AP}_{50:95}$ / planning ms / total ms.}
\label{tab:timing}
\end{table}

\paragraph{Latency breakdown and beam-width scaling.}
Table~\ref{tab:timing} shows that extending search from $H=1$ to $H=3$ adds just $5.1$~ms of planning on RUOD ($6.1$ to $11.2$~ms) and $4.7$~ms on DUO ($7.0$ to $11.7$~ms), keeping total episode latency within $67$--$80$~ms. Across $K\in\{4,8,12,24\}$ at $B=2$ and $H=3$, AP stays nearly constant ($0.3921$--$0.3923$ on DUO; $0.4013$--$0.4043$ on RUOD, with $K=8$ within $0.17$~pp of $K=24$), confirming robustness under symbolic pruning. Because fixed GPU kernel-launch and scheduling costs dominate batched transitions, narrowing $K$ from $24$ to $4$ trims planning modestly from $11.2$ to $10.6$~ms on RUOD and $11.7$ to $11.2$~ms on DUO.

\paragraph{Prediction fidelity and search horizon.}
Across the 32 dataset--budget combinations, $H=3$ improves over $H=1$ in 12 settings, while $H=5$ matches $H=3$ at $B=2$ where two fast queries exhaust the budget. Open-loop rollout diagnostics on held-out trajectories (detailed in the supplement) show that multi-step lookahead is most effective when the learned transition model maintains low state-prediction error relative to persistence: at $B=8$, five-step normalized root-mean-square error (RMSE) ratios are $0.680$ on RUOD and $0.710$ on DUO, compared with $1.054$ on TACO-4 and $1.045$ on TrashCan, where one-step policies suffice.

\section{Discussion and Limitations}
Our study reveals three structural boundaries of neuro-symbolic observation planning and practical limitations. First, while calibrated tariffs and serial GPU timings quantify performance on curated datasets, the planner has not yet been tested onboard physical robotic hardware (e.g., AUV/UAV) under live sensor streams, lighting or turbidity shifts, and thermal throttling. Second, our formulation models target-conditioned tool access with per-query costs; when a native single-pass detector extracts all classes at full view, planning shifts primarily to scheduling localized crops and enhancements ($45.2\%$--$83.4\%$ edited views at $B=8$). Third, although symbolic masks preserve feasibility and Dynamic curbs premature stopping, isolating slack pricing from fallback---and relaxing the eight-step cap while aligning F1 rewards with COCO AP---remains open.

\section{Conclusion}
This work casts budgeted visual inspection as neuro-symbolic observation planning under partial observability, decoupling compact learned perceptual dynamics from exact symbolic legality and latency-calibrated tariffs. Coupling multi-step lookahead with budget-aware pricing shows how explicit test-time search overcomes delayed-reward viewport traps, tariff cliffs, and conservative high-budget stopping across terrestrial, optical, and sonar domains without retraining frozen detectors.

Future work will pursue three directions: (i)~deploying the controller onboard embedded AUV/UAV hardware under live sensor latency and power limits; (ii)~training multi-step counterfactual reward and epistemic uncertainty heads for constrained UCT search; and (iii)~separating adaptive dual pricing from fallback while supporting variable episode horizons for many-class inspection.
\label{page:body-end}

%% file: sections/algorithm.tex
\begin{algorithm}[t]
\caption{Neuro-symbolic planning and control}
\label{alg:inspection}
\small
\begin{tabbing}
\hspace{0.9em}\=\hspace{0.9em}\=\hspace{0.9em}\=\hspace{0.9em}\=\kill
\textbf{Procedure} $\mathrm{Legal}(q=(b,v,\mathcal{H},n,\eta),\, a)$:\\
\> \textbf{if} $a=\stopact$ \textbf{then return} $\mathrm{true}$\\
\> \textbf{if} $n\ge T$ \textbf{or} $\kappa(a)>b$ \textbf{then return} $\mathrm{false}$\\
\> \textbf{if} $a\in\calA_{\mathrm{query}}$ \textbf{then return} $(a,v)\notin\mathcal{H}$\\
\> \textbf{return} $(\eta<3)\land\mathrm{GeomAndEditValid}(v,a)$\\[\smallskipamount]
\textbf{Procedure} $\mathrm{StepControl}(q=(b,v,\mathcal{H},n,\eta),\, a)$:\\
\> \textbf{if} $a\in\calA_{\mathrm{query}}$ \textbf{then return} $(b-\kappa(a),v,\mathcal{H}\cup\{(a,v)\},n+1,0)$\\
\> \textbf{return} $(b-\kappa(a),\,\mathrm{ApplyEdit}(v,a),\,\mathcal{H},\,n+1,\,\eta+1)$\\[\smallskipamount]
\textbf{Procedure} $\mathrm{Next}(\pi,q)$:\\
\> \textbf{if} $\pi$ ends in $\stopact$ \textbf{then return} $\{\stopact\}$\\
\> \textbf{return} $\{a\in\calA:\mathrm{Legal}(q,a)\}$\\[\smallskipamount]
\textbf{Main Loop} $\mathrm{PlanAndInspect}(I,\calC,B)$:\\
\> $(o,q,\calD)\gets\mathrm{Initialize}(I,\calC,B)$,\quad $h\gets[o]$\\
\> \textbf{while} episode is open \textbf{do}\\
\>\> $\lambda_t\gets\mathrm{Price}(q)$;\quad $z_t\gets e_\theta(h)$\\
\>\> $\mathcal P\gets\{(\epsilon,z_t,q,0)\}$\\
\>\> \textbf{for} $d=0,\ldots,\min(H,T-n)-1$ \textbf{do}\\
\>\>\> $\mathcal E\gets\emptyset$\\
\>\>\> \textbf{for} $(\pi,\hat z,\hat q,\widehat G)\in\mathcal P$ \textbf{and} $a\in\mathrm{Next}(\pi,\hat q)$ \textbf{do}\\
\>\>\>\> $\tilde r\gets \widehat r_\theta(\hat z,a)\cdot\mathbb{I}[a\in\calA_{\mathrm{query}}]$\\
\>\>\>\> $\widehat G'\gets\widehat G+\gamma^d[\tilde r-\lambda_t\kappa(a)]$\\
\>\>\>\> $\hat z'\gets f_\theta(\hat z,a)$ \textbf{if} $a\ne\stopact$ \textbf{else} $\hat z$\\
\>\>\>\> $\hat q'\gets \mathrm{StepControl}(\hat q,a)$ \textbf{if} $a\ne\stopact$ \textbf{else} $\hat q$\\
\>\>\>\> Add $(\pi a,\hat z',\hat q',\widehat G')$ to $\mathcal E$\\
\>\>\> $\mathcal P\gets$ top-$K$ prefixes in $\mathcal E$ by score $\widehat G'$\\
\>\> $(\pi^*,\cdot,\cdot,\widehat G^*)\gets\mathrm{Best}(\mathcal P)$;\quad $a_t\gets\pi^*_0$ \textbf{if} $\widehat G^*>0$ \textbf{else} $\stopact$\\
\>\> \textbf{if} fallback enabled, $a_t=\stopact$, $b\ge1$, $\mathcal U(q)\ne\emptyset$ \textbf{then}\\
\>\>\> $a_t\gets\arg\max_{a\in\mathcal U(q)}\widehat r_\theta(z_t,a)$\\
\>\> $(o,q,\calD)\gets\mathrm{Execute}(a_t)$;\quad $h\gets\mathrm{SlideWindow}(h,a_t,o)$\\
\> \textbf{return} $\calD$
\end{tabbing}
\end{algorithm}

%% file: tables/budget2_compact.tex
\begin{tabular*}{\textwidth}{@{\extracolsep{\fill}}lcccccccc@{}}
\toprule
 & \multicolumn{8}{c}{Test $\text{AP}_{50:95}$ ($B=2$)} \\
\cmidrule(lr){2-9}
 & \multicolumn{2}{c}{Terrestrial} & \multicolumn{3}{c}{Underwater Optical} & \multicolumn{3}{c}{Underwater Sonar} \\
\cmidrule(lr){2-3}\cmidrule(lr){4-6}\cmidrule(lr){7-9}
Method & COCO-5 & TACO-4 & DUO & RUOD & TrashCan & ARIS & SCTD & UATD \\
\midrule
Round robin & $0.177$ & $0.155$ & $0.298$ & $0.112$ & $0.030$ & $0.166$ & $0.092$ & $0.059$ \\
Enhance--query & $0.107$ & $0.000$ & $0.060$ & $0.055$ & $0.030$ & $0.092$ & $0.006$ & $0.040$ \\
Cascade & $0.177$ & $0.155$ & $0.298$ & $0.112$ & $0.030$ & $0.166$ & $0.092$ & $0.059$ \\
Fixed schedule & $0.177$ & $0.155$ & $0.298$ & $0.112$ & $0.030$ & $0.166$ & $0.092$ & $0.059$ \\
Random & $0.050$ & $0.010$ & $0.084$ & $0.066$ & $0.009$ & $0.049$ & $0.044$ & $0.021$ \\
Fixed priority* & $0.181^{*}$ & $\mathbf{0.172^{*}}$ & $0.324^{*}$ & $0.120$ & $\mathbf{0.033}$ & $0.076$ & $0.159$ & $0.065$ \\
\midrule
Imitation & $0.085{\pm}0.012$ & $0.000{\pm}0.000$ & $0.297{\pm}0.000$ & $0.093{\pm}0.005$ & $0.030{\pm}0.000$ & $0.097{\pm}0.009$ & $0.032{\pm}0.032$ & $0.040{\pm}0.000$ \\
CQL & $0.154{\pm}0.040$ & $0.155{\pm}0.000$ & $0.298{\pm}0.000$ & $0.100{\pm}0.019$ & $0.030{\pm}0.000$ & $0.166{\pm}0.000$ & $0.090{\pm}0.004$ & $0.047{\pm}0.010$ \\
Reward-only & $\mathbf{0.188{\pm}0.009}$ & $0.057{\pm}0.035$ & $0.365{\pm}0.005$ & $0.320{\pm}0.023$ & $0.027{\pm}0.001$ & $0.243{\pm}0.057$ & $0.158{\pm}0.029$ & $0.064{\pm}0.029$ \\
\midrule
WM ($H=1$) & $0.179{\pm}0.013$ & $0.151{\pm}0.013$ & $0.350{\pm}0.003$ & $0.345{\pm}0.033$ & $0.025{\pm}0.003$ & $\mathbf{0.279{\pm}0.054}$ & $0.198{\pm}0.033$ & $\mathbf{0.075{\pm}0.013}$ \\
WM ($H=3$) & $0.168{\pm}0.020$ & $0.159{\pm}0.011$ & $\mathbf{0.392{\pm}0.016}$ & $\mathbf{0.404{\pm}0.024}$ & $0.024{\pm}0.002$ & $0.228{\pm}0.038$ & $\mathbf{0.221{\pm}0.001}$ & $0.068{\pm}0.015$ \\
WM ($H=5$) & $0.168{\pm}0.020$ & $0.159{\pm}0.011$ & $\mathbf{0.392{\pm}0.016}$ & $\mathbf{0.404{\pm}0.024}$ & $0.024{\pm}0.002$ & $0.228{\pm}0.038$ & $\mathbf{0.221{\pm}0.001}$ & $0.068{\pm}0.015$ \\
\bottomrule
\end{tabular*}

%% file: tables/query_behavior.tex
\begin{tabular*}{\textwidth}{@{\extracolsep{\fill}}lcccccccc@{}}
\toprule
 & \multicolumn{2}{c}{Terrestrial} & \multicolumn{3}{c}{Underwater Optical} & \multicolumn{3}{c}{Underwater Sonar} \\
\cmidrule(lr){2-3}\cmidrule(lr){4-6}\cmidrule(lr){7-9}
Method / Metric & COCO-5 & TACO-4 & DUO & RUOD & TrashCan & ARIS & SCTD & UATD \\
\midrule
\multicolumn{9}{@{}l}{\emph{Tight budget ($B=2$)}}\\
WM ($H=1$) scene lift $\Delta_{\rm match}$ (pp) & $+3.8{\pm}0.1$ & $+1.1{\pm}1.5$ & $+4.8{\pm}1.0$ & $+37.3{\pm}1.5$ & $+16.9{\pm}7.1$ & $+33.9{\pm}1.6$ & $+3.1{\pm}3.5$ & $+11.3{\pm}2.9$ \\
WM ($H=3$) scene lift $\Delta_{\rm match}$ (pp) & $+3.0{\pm}0.7$ & $+1.0{\pm}1.1$ & $+6.1{\pm}1.4$ & $+34.3{\pm}2.4$ & $+18.0{\pm}6.0$ & $+29.0{\pm}3.2$ & $+3.0{\pm}1.5$ & $+10.9{\pm}4.3$ \\
WM ($H=3$ vs.\ $H=1$) set changed (\%) & $7.9$ & $16.6$ & $18.3$ & $39.9$ & $19.2$ & $51.4$ & $14.2$ & $30.5$ \\
WM ($H=3$) edited-view query (\%) & $1.3$ & $0.3$ & $0.0$ & $0.1$ & $8.4$ & $19.6$ & $0.6$ & $12.7$ \\
\midrule
\multicolumn{9}{@{}l}{\emph{Large budget ($B=8$, $H=3$)}}\\
WM ($H=3$) edited-view query (\%) & $72.2$ & $53.1$ & $83.4$ & $45.2$ & $36.6$ & $63.4$ & $77.8$ & $57.6$ \\
WM ($H=3$) present-class coverage (\%) & $73.4$ & $56.2$ & $91.9$ & $69.3$ & $19.6$ & $62.4$ & $85.8$ & $40.8$ \\
WM ($H=3$) stop reserve $\ge 1$ unit (\%) & $50.0$ & $64.4$ & $27.8$ & $65.7$ & $86.5$ & $48.1$ & $45.1$ & $73.9$ \\
\midrule
\multicolumn{9}{@{}l}{\emph{Dynamic pricing + legal fast fallback ($B=8$, $H=3$)}}\\
Dynamic present-class coverage (\%) & $86.9$ & $88.8$ & $93.9$ & $78.4$ & $54.1$ & $75.7$ & $90.7$ & $57.5$ \\
Dynamic stop reserve $\ge 1$ unit (\%) & $0.9$ & $7.2$ & $1.5$ & $0.0$ & $0.0$ & $0.0$ & $8.0$ & $0.0$ \\
Dynamic $\Delta\text{AP}_{50:95}$ vs.\ fixed (pp) & $+6.7$ & $+5.8$ & $+0.6$ & $+2.6$ & $+1.8$ & $+5.9$ & $+2.4$ & $+2.7$ \\
\bottomrule
\end{tabular*}

%% file: tables/timing_breakdown.tex
\begin{tabular*}{\columnwidth}{@{\extracolsep{\fill}}lrrrr@{}}
\toprule
 & \multicolumn{2}{c}{RUOD ($B=2$)} & \multicolumn{2}{c}{DUO ($B=2$)} \\
\cmidrule(lr){2-3}\cmidrule(lr){4-5}
Runtime stage (ms) & $H=1$ & $H=3$ & $H=1$ & $H=3$ \\
\midrule
Planning \& search & 6.1 & 11.2 & 7.0 & 11.7 \\
Tool \& state step & 20.8 & 23.4 & 30.2 & 32.6 \\
Diagnostic logging & 1.9 & 2.2 & 2.1 & 2.5 \\
Other overhead & 28.1 & 30.2 & 33.2 & 33.4 \\
\midrule
Total episode time & 56.8 & 67.0 & 72.5 & 80.2 \\
\bottomrule
\end{tabular*}
\par\smallskip
\begin{tabular*}{\columnwidth}{@{\extracolsep{\fill}}lcc@{}}
\toprule
 & RUOD ($\text{AP}_{50:95}$ / ms) & DUO ($\text{AP}_{50:95}$ / ms) \\
\cmidrule(lr){2-2}\cmidrule(lr){3-3}
$K$ ($H=3$) & AP / search / total & AP / search / total \\
\midrule
4 & 0.4013 / 10.6 / 64.3 & 0.3921 / 11.2 / 79.1 \\
8 & 0.4024 / 10.7 / 64.5 & 0.3923 / 11.4 / 78.5 \\
12 & 0.4043 / 10.9 / 63.1 & 0.3923 / 11.6 / 78.9 \\
24 & 0.4041 / 11.2 / 67.0 & 0.3923 / 11.7 / 80.2 \\
\bottomrule
\end{tabular*}

%% file: sections/ethics.tex
\section*{Ethical and Societal Impact}
This study evaluates budgeted visual inspection on publicly established terrestrial, underwater optical, and sonar datasets. Directing sensing effort toward scene-relevant targets under hardware or latency constraints supports marine environmental monitoring, autonomous underwater inspection, and assistive perception. Because cost-regularized termination balances expected detection gain against sensing expenditure, safety-critical deployments that require exhaustive absence verification should pair learned lookahead with explicit minimum-coverage guarantees and domain-specific validation protocols. All visual figures credit their benchmark sources, and ground-truth annotations are used strictly offline for training supervision and held-out evaluation.